%% file: paper.tex
\documentclass[a4paper,twoside]{article}

\usepackage{epsfig}
\usepackage{subcaption} 
\usepackage{calc}
\usepackage{amssymb}
\usepackage{amstext}
\usepackage{amsmath}
\usepackage{amsthm}
\usepackage{multicol}
\usepackage{pslatex}
\usepackage{apalike}
\usepackage[bottom]{footmisc}
\usepackage{SCITEPRESS} 

\usepackage{cite}
\usepackage{float}
\usepackage{siunitx} 
\usepackage{booktabs} 
\usepackage{geometry} 
\usepackage{graphicx} 
\usepackage{textcomp}
\usepackage{xcolor} 
\usepackage{xspace}
\usepackage[linesnumbered,ruled,vlined]{algorithm2e}
\usepackage[hyphens]{url} 

\usepackage{multirow} 

\begin{document}

\title{Enhanced Intrusion Detection in IIoT Networks: A Lightweight Approach with Autoencoder-Based Feature Learning}

\author{
    \authorname{Tasnimul Hasan\sup{1}, Abrar Hossain\sup{1}, Mufakir Qamar Ansari\sup{1}, and Talha Hussain Syed\sup{1}}
    \affiliation{\sup{1}Department of Electrical Engineering and Computer Science, The University of Toledo, Toledo, OH, USA}
    \email{\{tasnimul.hasan, abrar.hossain, mufakir.ansari, talhahussain.syed\}@utoledo.edu}
}


\abstract{
The rapid expansion of the Industrial Internet of Things (IIoT) has significantly advanced digital technologies and interconnected industrial sys- tems, creating substantial opportunities for growth. However, this growth has also heightened the risk of cyberattacks, necessitating robust security measures to protect IIoT networks. Intrusion Detection Systems (IDS) are essential for identifying and preventing abnormal network behaviors and malicious activities. Despite the potential of Machine Learning (ML)-based IDS solutions, existing models often face challenges with class imbalance and multiclass IIoT datasets, resulting in reduced detection accuracy. This research directly addresses these challenges by implementing six innovative approaches to enhance IDS perfor- mance, including leveraging an autoencoder for di- mensional reduction, which improves feature learning and overall detection accuracy. Our proposed Decision Tree model achieved an exceptional F1 score and accuracy of 99.94\% on the Edge-IIoTset dataset. Furthermore, we prioritized lightweight model design, ensuring deployability on resource-constrained edge devices. Notably, we are the first to deploy our model on a Jetson Nano, achieving inference times of 0.185 ms for binary classification and 0.187 ms for multiclass classification. These results highlight the novelty and robustness of our approach, offering a practical and efficient solution to the challenges posed by imbalanced and multiclass IIoT datasets, thereby enhancing the detection and prevention of network intrusions.
}

\onecolumn \maketitle \normalsize \setcounter{footnote}{0} \vfill

\input{01_intro}

\input{02_related_work}

\input{03_system_architecture}

\input{04_performance_eval}
\input{05_conclusion}

\bibliographystyle{apalike}
{\small
\bibliography{ref}}

\vspace{12pt}

\end{document}

%% file: 01_intro.tex
\section{Introduction}
The Industrial Internet of Things (IIoT) integrates traditional industrial processes with advanced digital technologies, enabling the seamless interconnection of sensors, devices, and systems across various sectors\cite{lynn2020internet}. This integration enhances operational efficiency, productivity, and innovation through real-time data collection, analysis, and decision-making. IIoT is applied in industries such as manufacturing, energy, healthcare, and transportation, where it supports predictive maintenance, asset tracking, remote monitoring, and smart automation\cite{xu2018industry}. However, the extensive interconnectivity in IIoT systems makes them vulnerable to cyber threats, exacerbated by weak authentication mechanisms and irregular security updates\cite{alani2023explainable}. The vast number of connected devices and diverse communication protocols create multiple entry points for cybercriminals\cite{hassan2019current}, exposing IIoT networks to malware, data breaches, denial-of-service (DoS) attacks, and advanced persistent threats (APTs)\cite{yugha2020survey}. These threats can lead to significant operational disruptions, financial losses, and compromise the safety of critical infrastructure\cite{hassija2019survey}.

Intrusion Detection Systems (IDSs) are essential for detecting and preventing unauthorized access and abnormal behavior in networks\cite{elrawy2018intrusion}. IDSs monitor network traffic in real time, identifying potential threats using methods like signature-based, anomaly-based, and hybrid detection\cite{buczak2015survey, leu2015internal, mirsky2018kitsune}. While these methods are effective, they face challenges in the complex and large-scale IIoT environments\cite{zarpelao2017survey}, highlighting the need for more advanced IDS solutions. Moreover, current IDSs perform well with a limited number of classes and balanced data\cite{karatas2020increasing}. However, real-world scenarios often involve more classes and significant data imbalance, where infrequent yet critical classes may be overlooked. This imbalance can degrade detection performance, increasing the risk of undetected security threats. To address these challenges, performing machine learning (ML) inference on edge devices is crucial. Edge inference allows real-time traffic analysis, reducing latency and reliance on centralized systems while enhancing privacy and security. Lightweight ML models designed for edge environments operate efficiently under resource constraints, enabling robust intrusion detection even in remote or bandwidth-limited IIoT settings\cite{hossain2025hpc}.

To address these challenges, we make the following contributions in this paper:
\begin{itemize}
\item We propose a novel intrusion detection system (IDS) that leverages an autoencoder for dimensionality reduction, effectively addressing challenges with imbalanced and multiclass IIoT datasets while enhancing feature learning.
\item We provide robust evidence of our IDS's effectiveness, achieving an exceptional F1 score and accuracy of 99.94
\item We highlight the practical applicability of our approach by being the first to deploy the top-performing IDS models on a Jetson Nano, achieving fast inference times of 0.185 ms for binary classification and 0.187 ms for multiclass classification, making it suitable for real-world edge computing environments.
\end{itemize}

%% file: 02_related_work.tex
\section{Related Work}\label{sec:related_work}

A notable approach in IoT/IIoT network security is the Lightweight Stacking Ensemble Learning (SEL) method combined with Feature Importance (FI) for dimensionality reduction. This method reduces storage requirements by 86.9\% while maintaining a high classification accuracy of 99.96\% \cite{abdulkareem2024lightweight}. SEL’s performance, particularly in multiclass classification scenarios, surpasses traditional models like Decision Trees and SVMs \cite{hassini2024end}. Similarly, a CNN1D model offers an end-to-end approach for intrusion detection, achieving 99.96\% accuracy across 15 attack classes \cite{saadouni2023secure}. Its strength lies in efficiently handling feature extraction and classification, making it valuable in real-time Industrial IoT environments Furthermore, a Self-Attention-based Deep Convolutional Neural Network (SA-DCNN) enhances feature prioritization, improving detection in IIoT networks with an accuracy of 99.95\% on the Edge-IIoTset \cite{alshehri2024self}. Deploying edge ML and HPC   

The Edge-IIoTset dataset stands out for its comprehensive coverage of IoT/IIoT devices and protocols, supporting both centralized and federated learning models. It is more versatile compared to MQTTset and WUSTL-IIoT-2021, which have narrower focuses \cite{ferrag2022edge}. Hybrid models like CNN-LSTM and CNN-GRU are also effective in IIoT security. The CNN-LSTM model excels in feature extraction and sequence prediction, particularly for complex attacks like MITM , while the CNN-GRU model optimizes time-series data classification, further enhancing IIoT network security \cite{saadouni2023secure}. Additionally, the Multi-Head Attention-based Gated Recurrent Unit (MAGRU) model addresses data imbalance and complex class structures in IIoT, achieving high accuracy and precision on datasets like Edge-IIoTset and MQTTset \cite{ullah2023magru}.

%% file: 03_system_architecture.tex
\section{Methodology}\label{sec:system_architecture}

This section outlines our methodology for designing an effective IDS for IIoT. Using the Edge-IIoTset dataset, we preprocess data through feature selection, normalization, and encoding. To tackle class imbalance, we introduce a cost-sensitive autoencoder that prioritizes underrepresented attack types. Finally, we implement a lightweight architecture optimized for deployment on resource-constrained edge devices, balancing accuracy and efficiency.

\subsection{Dataset}
The Edge-IIoTset dataset, comprising 2.2 million records from a seven-layer, ten-device setup, includes 1.6 million normal traffic instances and over 600,000 across 14 attack types, highlighting significant imbalance. Similarly, the MQTTset, with 541,000 records from eight sensors, also skews heavily toward normal traffic. Edge-IIoTset, featuring 61 novel detection-enhancing features, realistically represents imbalanced IIoT traffic, making it vital for developing robust intrusion detection systems.

\begin{table}[htbp]
\caption{Dataset Distribution of Edge-IIoTset}
\label{tab:edge-iiotset}
\centering 
\renewcommand{\arraystretch}{1.5} 
\begin{tabular}{|p{1.0cm}|p{2.2cm}|p{2.8cm}|}
\hline
 & \textbf{Traffic Class} & \textbf{Records} \\
\hline
\textbf{Normal} & Normal & {1,380,858 (71.65\%)} \\
\hline
\multirow{14}{*}{\textbf{Attack}} 
 & DDoS\_UDP & {121,567 (6.31\%)} \\
\cline{2-3}
 & DDoS\_ICMP & {67,939 (3.53\%)} \\
\cline{2-3}
 & SQL\_injection & {50,826 (2.64\%)} \\
\cline{2-3}
 & DDoS\_TCP & {50,062 (2.60\%)} \\
\cline{2-3}
 & Vulnerability & {50,026 (2.60\%)} \\
\cline{2-3}
 & Password & {49,933 (2.59\%)} \\
\cline{2-3}
 & DDoS\_HTTP & {49,203 (2.55\%)} \\
\cline{2-3}
 & Uploading & {36,915 (1.92\%)} \\
\cline{2-3}
 & Backdoor & {24,026 (1.25\%)} \\
\cline{2-3}
 & Port\_Scanning & {19,983 (1.04\%)} \\
\cline{2-3}
 & XSS & {15,066 (0.78\%)} \\
\cline{2-3}
 & Ransomware & {9,689 (0.50\%)} \\
\cline{2-3}
 & Fingerprinting & {853 (0.04\%)} \\
\cline{2-3}
 & MITM & {358 (0.02\%)} \\
\hline
\textbf{Total} &  & \textbf{1,927,304 (100\%)} \\
\hline
\end{tabular}
\end{table}

Table \ref{tab:edge-iiotset} shows the distribution of attack types in our dataset. Below is a brief summary of these attacks:

 \begin{itemize}
    \item \textbf{Normal (71.65\%)}: Legitimate network traffic without malicious activity.
    \item \textbf{DDoS UDP (6.31\%)}, \textbf{DDoS ICMP (3.53\%)}, \textbf{DDoS TCP (2.60\%)}, \textbf{DDoS HTTP (2.55\%)}: DDoS attacks using UDP, ICMP, TCP, or HTTP packets to overwhelm network resources.
    \item \textbf{SQL Injection (2.64\%)}: Inserting malicious SQL code to manipulate databases.
    \item \textbf{Vulnerability Scanner (2.60\%)}: Probing systems for security weaknesses.
    \item \textbf{Password (2.59\%)}: Attempting to crack or guess passwords.
    \item \textbf{Uploading (1.92\%)}: Unauthorized file uploads to compromise security.
    \item \textbf{Backdoor (1.25\%)}: Bypassing authentication for unauthorized access.
    \item \textbf{Port Scanning (1.04\%)}: Identifying open ports and services as a precursor to attacks.
    \item \textbf{XSS (0.78\%)}: Injecting malicious scripts into webpages (Cross-Site Scripting).
    \item \textbf{Ransomware (0.50\%)}: Encrypting data and demanding payment for decryption.
    \item \textbf{Fingerprinting (0.04\%)}: Gathering device information for targeted attacks.
    \item \textbf{MITM (0.02\%)}: Intercepting and altering communication between two parties (Man-in-the-Middle).
\end{itemize}

\begin{algorithm}[htbp]
\SetAlgoLined
\SetKwInOut{KwData}{Data}
\SetKwInOut{KwResult}{Result}
\caption{Training a Deep Autoencoder on Edge-IIoTset Dataset}
\label{alg:cost_sensitive_autoencoder}

\KwData{Training and validation data in DataFrames}
\KwResult{Trained autoencoder model}

Load training and validation samples into DataFrames\;
Initialize and compile the model with Adam optimizer and MSE loss\;
Define ModelCheckpoint for saving the best model and EarlyStopping to prevent overfitting\;
Fit the model on the training set and validate on the validation set\;
Save the best model\;

\end{algorithm}

\subsection{Data Preprocessing}
\subsubsection{Feature Selection} Irrelevant columns were removed as they provided no predictive value, while columns with constant values and highly correlated features\((\text{correlation} > 0.6)\)
 were dropped to reduce redundancy and computational complexity.

\subsubsection{Label Encoding} Categorical variables were converted to numerical format using label encoding to ensure compatibility with machine learning algorithms.

\subsubsection{Normalization} Features were normalized to a [0, 1] range using Min-Max Scaling, preventing scale dominance and improving training efficiency for sensitive models like neural networks.

\subsection{Addressing Class Imbalance with Cost-Sensitive Autoencoder} 
\begin{figure}[!]
    \centering
    \includegraphics [width=0.9\linewidth]{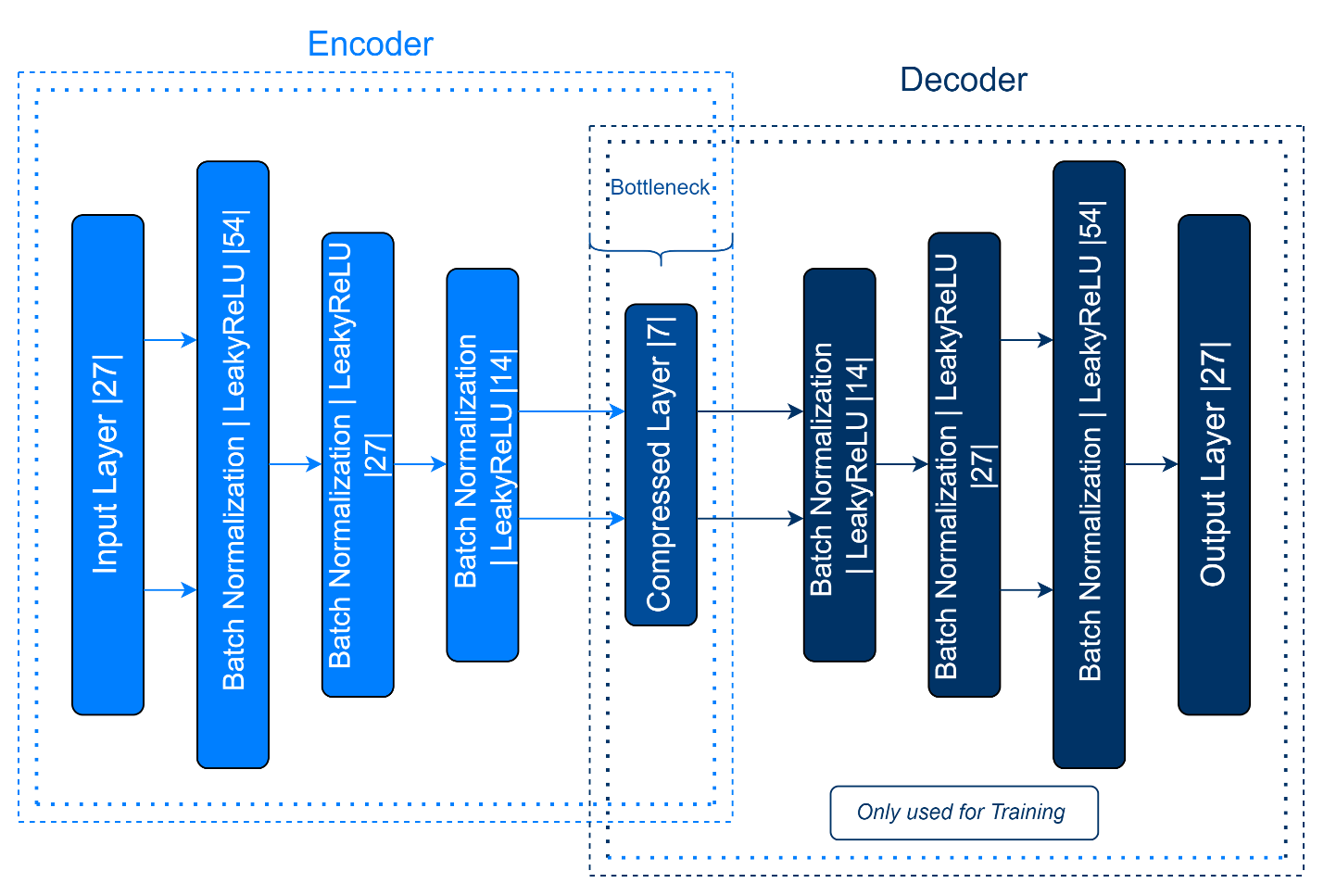}
    \caption{Architecture of the proposed cost-sensitive autoencoder.}
    \label{fig:autoencoder_block_diagram}
\end{figure}

As shown in Fig. \ref{fig:autoencoder_block_diagram} we introduce a cost-sensitive autoencoder to address the imbalanced nature of the Edge-IIoTset dataset.
An autoencoder is a type of artificial neural network designed to learn efficient data representations by encoding inputs into a compressed latent space and then reconstructing the output as closely as possible to the original input.

The proposed autoencoder sets itself apart from existing approaches, such as variational autoencoders (VAEs), by focusing on architectural simplicity and addressing data imbalance directly. While VAEs aim to model data distributions through probabilistic latent representations, our approach is deterministic, prioritizing compact and discriminative feature learning for accurate reconstruction. By excluding the probabilistic layers used in VAEs, our autoencoder reduces computational complexity and is better suited for resource-constrained environments, such as edge devices. This deterministic approach also ensures stability in reconstruction, particularly when handling highly imbalanced datasets like Edge-IIoTset. Architecturally, our autoencoder is designed to efficiently handle the specific challenges posed by IIoT datasets. The encoder compresses the input feature space from 24 dimensions to a bottleneck layer of 6 dimensions, retaining critical features while discarding redundant information. This reduction is mathematically represented as:

\[
h = f(x) = \sigma(Wx + b)
\]

where \( W \) and \( b \) denote the weights and biases of the encoder, and \( \sigma \) is the nonlinear activation function. This compressed latent representation, \( h \), captures the essential patterns of the input data. The decoder then reconstructs the input from \( h \), expanding it back to the original dimensionality, as given by:

\[
\hat{x} = g(h) = \sigma(W'h + b')
\]

where \( W' \) and \( b' \) represent the weights and biases of the decoder. The reconstruction objective is to minimize the difference between the original input \( x \) and the reconstructed output \( \hat{x} \). This difference is typically quantified using the mean squared error (MSE) loss function, defined as:

\[
L = \frac{1}{N} \sum_{i=1}^{N} w_{y_i} (x_i - \hat{x}_i)^2
\]

Here, \( N \) represents the number of inputs, \( x_i \) and \( \hat{x}_i \) are the original and reconstructed values, and \( w_{y_i} \) denotes the class weight for the \( i \)-th instance.

To address the pronounced class imbalance in the Edge-IIoTset dataset, we employ a cost-sensitive learning mechanism using class weights. These weights are assigned inversely proportional to the frequency of each class, ensuring that minority classes, such as "MITM" and "Fingerprinting," receive greater attention during training. By amplifying the contribution of these underrepresented classes to the loss function, the autoencoder becomes more sensitive to detecting subtle variations in their patterns. This integration of class weighting into the training process eliminates the need for external methods like oversampling or data augmentation, which can introduce biases or increase computational costs. The results demonstrate the effectiveness of this strategy, as the model achieves superior performance on rare attack types without compromising overall accuracy. Compared to traditional autoencoders or VAEs, our method is more efficient, scalable, and specifically tailored for real-world deployment in IIoT environments. The combination of architectural optimization and cost-sensitive learning underscores the novelty of our approach and its ability to address the unique challenges of intrusion detection in imbalanced IIoT datasets. 
\begin{figure}[!]
    \centering
    \includegraphics [width=1\linewidth]{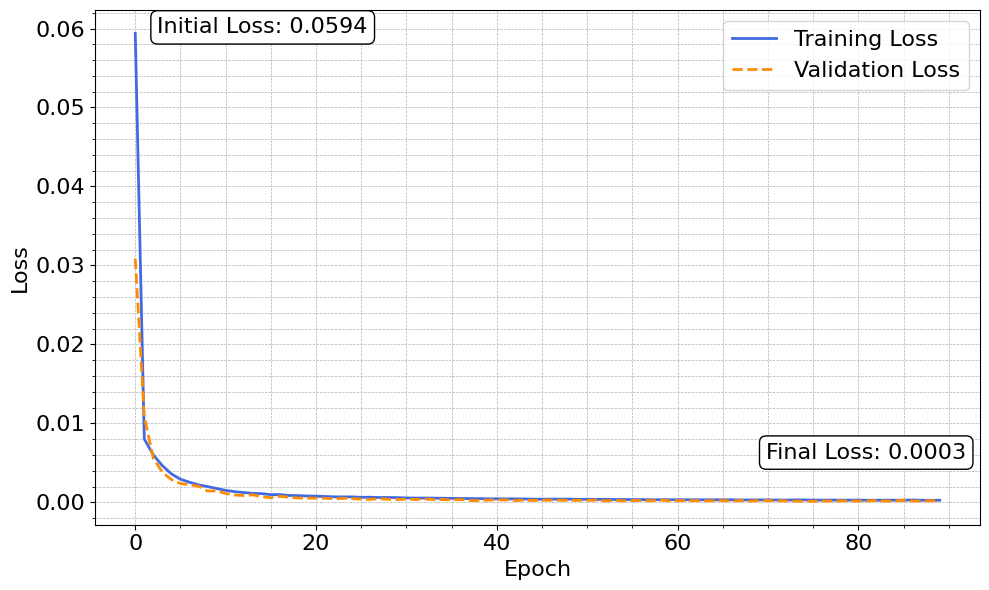}
    \caption{Autoencoder Loss Curve}
    \label{fig:autoencoder_loss_curve}
\end{figure}

\begin{figure}[!]
    \centering
    \includegraphics [width=1\linewidth]{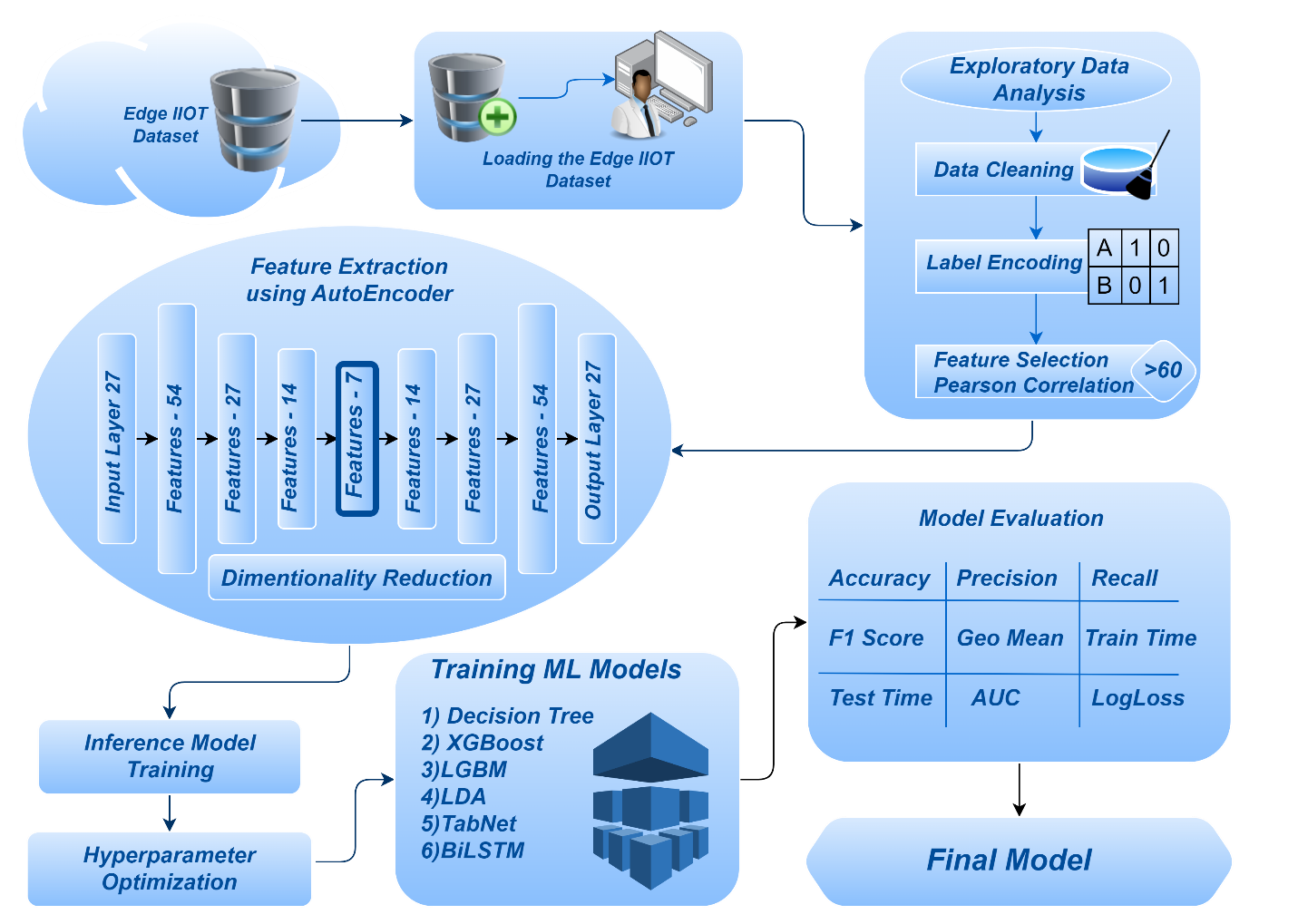}
    \caption{Proposed AutoEncoder based IDS Module}
    \label{fig:block_diagram}
\end{figure}
Fig. \ref{fig:autoencoder_loss_curve} shows the training and validation loss curves of our autoencoder model which employs class weights inversely proportional to their frequencies, emphasizing the learning of underrepresented attack types. As shown in Fig. 4 this approach enhances the model's ability to detect less frequent but critical attack types within the dataset, effectively addressing the class imbalance issue and improving the overall robustness of the intrusion detection system.

%% file: 04_performance_eval.tex
\section{Performance Evaluation}\label{sec:performance_eval}
In this section, we evaluate the performance of our proposed approach.

\subsection{Experimental Setup}

The experiments were conducted on two platforms: a workstation with Ubuntu 20.04.3 LTS, 16 GB RAM, and a 2.20 GHz Intel Xeon processor, and a Jetson Nano with a 128-core Maxwell GPU and a 1.43 GHz Quad-core ARM A57 CPU. The workstation used Python 3.10.12, while the Jetson Nano used Python 3.7. 

\subsection{Evaluation metrics}

The performance evaluation of various models used is summarized in the following tables and the confusion matrix. We evaluated our models based on the  following metrics.

\begin{equation}\label{eq:accuracy}
\scriptsize
Accuracy = \frac{TP+TN}{TP+FP+TN+FN}   
 \end{equation}
\begin{equation}\label{eq:precision}
\scriptsize
Precision = \frac{TP}{TP+FP} \qquad 
\end{equation}
\begin{equation}\label{eq:recall}
\scriptsize
Recall = \frac{TP}{TP+FN}  \qquad 
\end{equation}
\begin{equation}\label{eq:f1_eqn}
\scriptsize
    F1 = 2 * \frac{Precision*Recall}{Precision + Recall}              
\end{equation}

\begin{figure*}[t!]
\centering
\begin{subfigure}{0.45\textwidth}
    \includegraphics[width=\textwidth]{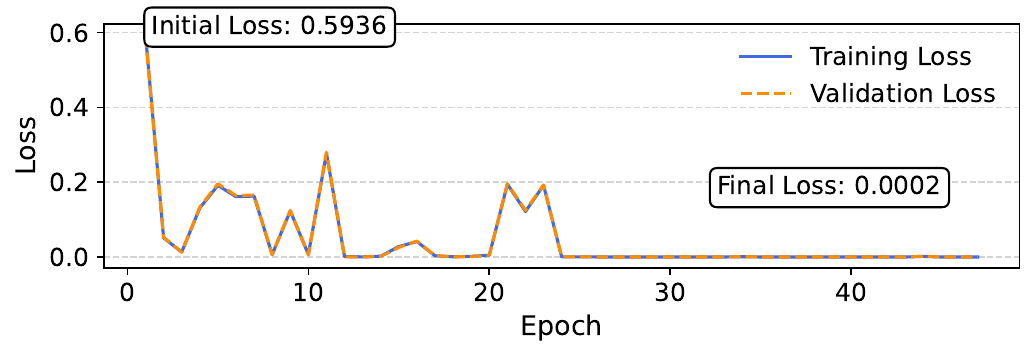}
    \caption{Tabnet/Binary Loss curve}
\end{subfigure}
\hspace{0.5cm}
\begin{subfigure}{0.45\textwidth}
    \includegraphics[width=\textwidth]{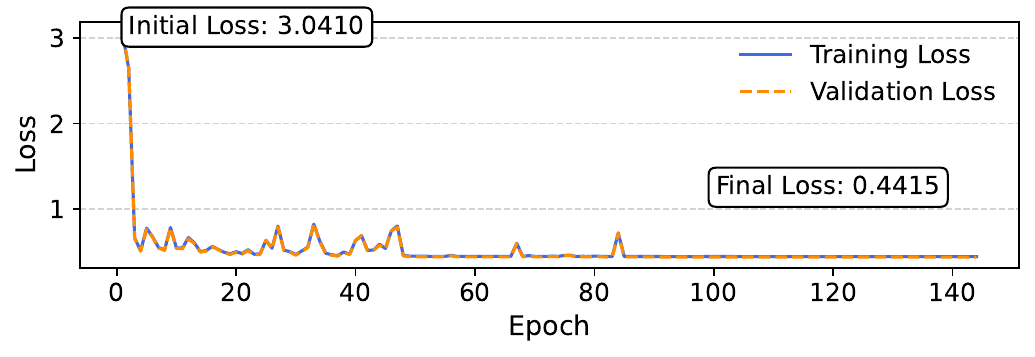}
    \caption{Tabnet/Multi Loss curve}
\end{subfigure}

\vspace{0.5cm}

\begin{subfigure}{0.45\textwidth}
    \includegraphics[width=\textwidth]{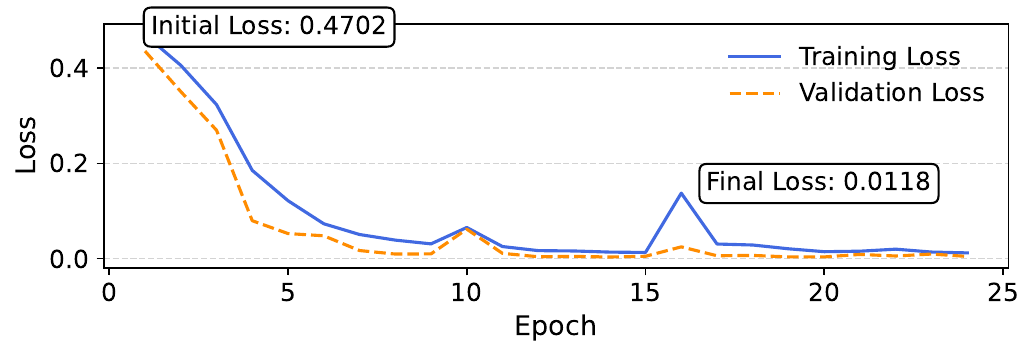}
    \caption{BiLSTM/Binary Loss curve}
\end{subfigure}
\hspace{0.5cm}
\begin{subfigure}{0.45\textwidth}
    \includegraphics[width=\textwidth]{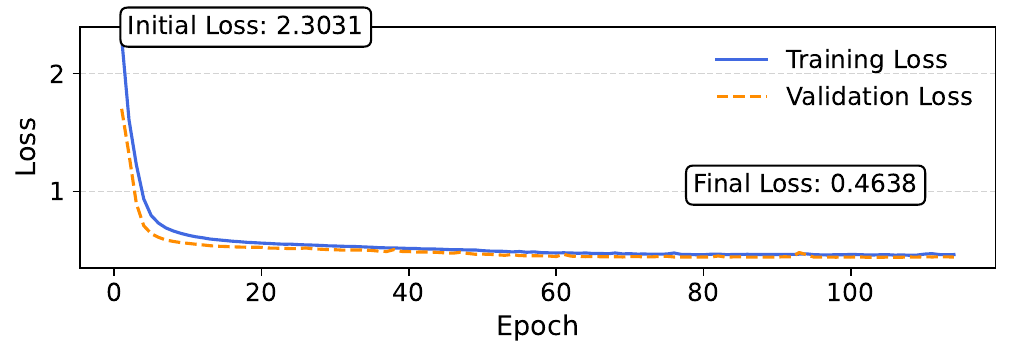}
    \caption{BiLSTM/Multi Loss curve}
\end{subfigure}

\caption{Training and Validation Loss for Tabnet and BiLSTM for Binary and Multiclass Classification}
\label{training_validation_losses}
\end{figure*}

\subsection{Confusion Matrix}

Table \ref{tab:confusion_matrix}  presents the confusion matrix for several models, with the LGBM model achieving perfect detection accuracy (4820 TPs, 25620 TNs, 0 FP, 0 FN), while the XGB model also performs well (4818 TPs, 25619 TNs, 2 FPs, 1 FN). In contrast, the LDA model shows significantly weaker performance (1771 TPs, 24876 TNs, 3049 FPs, 744 FNs).

\begin{table}[htbp]
\centering
\caption{Confusion Matrix for Various Models (Absolute and Normalized Percentage Values)}
\label{tab:confusion_matrix}
\scriptsize

\begin{tabular}{|>{\raggedright\arraybackslash}p{1.0cm}|>{\centering\arraybackslash}p{1.cm}|>{\centering\arraybackslash}p{1.cm}|>{\centering\arraybackslash}p{1.0cm}|>{\centering\arraybackslash}p{1.0cm}|}
\hline
\textbf{Model} & \textbf{Pred.\ Normal (TP)} & \textbf{Pred.\ Attack (FP)} & \textbf{Pred.\ Normal (FN)} & \textbf{Pred.\ Attack (TN)} \\ \hline
XGB & 4818 (15.83\%) & 2 (0.01\%) & 1 (0.00\%) & 25619 (84.16\%) \\ \hline
LGBM & 4820 (15.83\%) & 0 (0.00\%) & 0 (0.00\%) & 25620 (84.17\%) \\ \hline
LDA & 1771 (5.82\%) & 3049 (10.02\%) & 744 (2.44\%) & 24876 (81.72\%) \\ \hline
DT & 4817 (15.82\%) & 3 (0.01\%) & 2 (0.01\%) & 25618 (84.16\%) \\ \hline
TabNet & 4816 (15.86\%) & 3 (0.01\%) & 0 (0.00\%) & 25545 (84.13\%) \\ \hline
BiLSTM & 4812 (15.81\%) & 8 (0.03\%) & 0 (0.00\%) & 25620 (84.17\%) \\ \hline
\end{tabular}
\end{table}

\begin{table*}[]
\centering
\caption{Summary of model performance (Binary)}
\label{tab:binary_model_performance}
\begin{tabular}{l p{1.2cm} p{0.8cm} p{0.8cm} p{0.8cm} p{0.8cm} p{1.2cm} p{1.8cm} p{1.8cm}}
\toprule
Algorithm & Accuracy & Precision & Recall  & F1  & AUC & Geo Mean & Train Time (s) & Test Time (s) \\
\midrule
LGBM         & 1.0000 & 1.0000 & 1.0000 & 1.0000 & 1.0000 & 1.0000 & 1.0752   & 0.1099 \\
XGB          & 0.9999 & 0.9999 & 1.0000 & 0.9999 & 0.9996 & 0.9998 & 1.0283   & 0.0597 \\
DecisionTree & 0.9998 & 0.9999 & 0.9999 & 0.9999 & 0.9994 & 0.9996 & 1.1718   & 0.0042 \\
LDA          & 0.8754 & 0.8908 & 0.9710 & 0.9292 & 0.4487 & 0.5973 & 0.1600   & 0.0072 \\
TabNet       & 0.9999 & 0.9999 & 1.0000 & 0.9999 & 1.0000 & 0.9997 & 1735.4461 & 2.7580 \\
Bi-LSTM      & 0.9997 & 0.9997 & 1.0000 & 0.9998 & 0.9998 & 0.9920 & 0.9990 & 982.2894 \\
\bottomrule
\end{tabular}
\end{table*}

\begin{table*}[t]
\centering
\caption{Summary of model performance (Multi Class)}
\label{tab:multi_model_performance}
\begin{tabular}{l p{1.0 cm} p{0.6cm} p{0.8cm} p{0.8cm} p{0.8cm} p{0.8cm} p{0.8cm} p{0.8cm} p{0.8cm} p{1.0cm}}
\toprule
Algorithm & Accuracy & \multicolumn{2}{p{2.2cm}}{Precision} & \multicolumn{2}{p{2.2cm}}{Recall} & \multicolumn{2}{p{2.2cm}}{F1 Score} & Geo Mean & Train Time(s) & Test Time(s) \\
\cmidrule(lr){3-4} \cmidrule(lr){5-6} \cmidrule(lr){7-8}
 &  & Macro & Weighted & Macro & Weighted & Macro & Weighted &  &  &  \\
\midrule
DecisionTree & 0.9994 & 0.9994 & 0.9994 & 0.9994 & 0.9994 & 0.9994 & 0.9994 & 0.9997 & 3.7432 & 0.0124 \\
XGB & 0.8181 & 0.8106 & 0.8233 & 0.7974 & 0.8181 & 0.801 & 0.8183 & 0.8872 & 33.9725 & 0.8476 \\
LGBM & 0.7946 & 0.7915 & 0.8058 & 0.7737 & 0.7946 & 0.7768 & 0.795 & 0.8731 & 18.5499 & 4.8402 \\
LDA & 0.4663 & 0.3762 & 0.4359 & 0.3677 & 0.4663 & 0.3333 & 0.4261 & 0.5946 & 0.2051 & 0.0178 \\
TabNet & 0.7756 & 0.7751 & 0.7905 & 0.7488 & 0.7756 & 0.7548 & 0.7769 & 0.8584 & 4436.65 & --- \\
LSTM & 0.7729 & 0.7682 & 0.7882 & 0.74 & 0.7729 & 0.7279 & 0.7567 & 0.8532 & 958.55 & 3.1581 \\
\bottomrule
\end{tabular}
\end{table*}

\subsection{Results of Binary Classification}
We present the results of  binary classification in Table \ref{tab:binary_model_performance}. LGBM achieves perfect performance across all metrics, while XGB and Decision Tree models also perform well but with slightly longer training times. In contrast, LDA shows significantly lower accuracy at 0.8754, indicating its inadequacy for this task. Although TabNet and Bi-LSTM deliver near-perfect scores, their high training times (1735.4461 seconds and 982.2894 seconds, respectively) suggest a trade-off between accuracy and computational efficiency.

\subsection{Results of Multi Class Classification}

 Table \ref{tab:multi_model_performance}
summarizes our multi-class classification results using DT, XGB, LGBM, LDA, TabNet, and LSTM models, achieving 99.94\% accuracy and F1 score, demonstrating strong anomaly detection capabilities. The Decision Tree model excels with near-perfect metrics and the fastest test time (0.0124 seconds), while LDA shows the lowest performance (0.4663 accuracy). XGB and LGBM achieve moderate accuracy but have longer training times. Although TabNet and LSTM perform well, their lengthy training times, particularly TabNet's 4436.6573 seconds, highlight the trade-offs.

\subsection{Edge Inference Results}

To demonstrate the lightweight nature of our model, we conducted inference tests using our best-performing models, Decision Tree and LGBM, on a Jetson Nano. For binary classification, the total inference time was 5.62 seconds, with an average time per instance of 0.184 ms. For multiclass classification, the total inference time was 5.70 seconds, with an average time per instance of 0.187 ms.

\subsection{Comparison with Other Works}

We compared the performance of our model against four recent intrusion detection techniques: \cite{alshehri2024self}, \cite{douiba2023improved}, \cite{ferrag2022edge}, and \cite{ullah2023magru}. While these methods exhibit promising results, they also face significant limitations, particularly in handling imbalanced datasets and multiclass classification scenarios, which are prevalent in IIoT environments. For instance, approaches such as \cite{alshehri2024self} and \cite{douiba2023improved} struggle with class imbalance, often prioritizing majority classes while neglecting minority classes. This results in lower F1-scores and reduces their effectiveness in capturing nuanced patterns of intrusion. On the other hand, models like \cite{ferrag2022edge} and \cite{ullah2023magru} achieve high accuracy through complex architectures but fail to address the computational constraints of edge devices, making them unsuitable for real-time IIoT applications. In contrast, our approach not only outperforms these techniques but also addresses these limitations comprehensively. By leveraging an autoencoder for dimensionality reduction and feature learning, our model effectively mitigates the challenges posed by class imbalance. This is evident from our superior F1-score, as shown in Table \ref{tab:multi_model_performance}, which demonstrates our ability to accurately classify both majority and minority classes. Additionally, while the accuracy metric is commonly used, it can be misleading in imbalanced settings; our model achieves an outstanding accuracy of 99.94\%, striking a balance between precision and robustness that outperforms existing methods. Furthermore, our approach is novel in its emphasis on practical deployment. Unlike prior work, our model introduces a lightweight architecture explicitly designed for edge environments, ensuring scalability and efficiency. We are the first to deploy an intrusion detection model on a Jetson Nano, achieving exceptionally low inference times of 0.185 ms for binary classification and 0.187 ms for multiclass classification. 
\begin{table}[htbp]
\vspace{-5pt}
\centering
\caption{Comparison of Techniques and Performance Metrics}
\label{tab:comparison}
\scriptsize
\begin{tabular}{|>{\raggedright\arraybackslash}p{1.4cm}|>{\centering\arraybackslash}p{1.6cm}|>{\centering\arraybackslash}p{0.8cm}|>{\centering\arraybackslash}p{0.6cm}|>{\centering\arraybackslash}p{0.4cm}|}
\hline
\textbf{Authors} & \textbf{Techniques} & \textbf{Accuracy} & \textbf{F1} & \textbf{Edge} \\ \hline
Alshehri et al. & SA-DCNN & 99.95\% & 99.53\% & No \\ \hline
Douiba et al. & GB \& DT & 100\% & 99.50\% & No \\ \hline
Ferrag et al. & DT, RF, SVM, KNN, DNN & 94.67 & 99\% & No \\ \hline
Ullah et al. & MAGRU & 99.97\% & 99.64\% & No \\ \hline
\textbf{Our Approach} & \textbf{DT, XGB, LGBM, LDA, TabNet, LSTM} & \textbf{99.94\%} & \textbf{99.94\%} & \textbf{Yes} \\ \hline
\end{tabular}
\vspace{-10pt}
\end{table}
This capability demonstrates our model's suitability for real-world IIoT applications, where low-latency and resource efficiency are paramount. These contributions collectively establish our method as a robust, high-performance, and deployable solution that addresses the critical challenges faced by state-of-the-art intrusion detection systems.

%% file: 05_conclusion.tex
\section{Conclusion}
\label{sec:conclude}
Our Intrusion Detection System (IDS) achieved a 99.94\% F1 score on the Edge-IIoT dataset, overcoming significant data imbalance. This success is attributed to our autoencoder-based methodology, which enhances feature learning and detection accuracy while being lightweight for edge inference, making it suitable for real-world IoT environments. Pretraining the autoencoder enabled efficient knowledge transfer, further boosting performance. Looking ahead, we plan to expand our research by testing the model on additional datasets and diverse IoT settings, integrating advanced machine learning algorithms, and refining feature extraction techniques to detect more sophisticated attacks. We also aim to improve the model’s robustness and scalability for real-time deployment in large-scale IoT networks.